\documentclass[letterpaper]{article} 
\usepackage{aaai25}  
\usepackage{times}  
\usepackage{amsmath}
\usepackage{helvet}  
\usepackage{courier}  
\usepackage[hyphens]{url}  
\usepackage{graphicx} 
\usepackage{natbib}  
\usepackage{caption} 
\usepackage{algorithm}
\usepackage{algorithmic}

\usepackage{newfloat}
\usepackage{amssymb}

\usepackage{listings}
\DeclareCaptionStyle{ruled}{labelfont=normalfont,labelsep=colon,strut=off} 
\floatstyle{ruled}
\newfloat{listing}{tb}{lst}{}
\floatname{listing}{Listing}
\title{Advancing Personalized Learning Analysis via an Innovative Domain Knowledge Informed Attention-based Knowledge Tracing Method}
\author{
    Shubham Kose,
    Jin Wei-Kocsis\\
}
\affiliations{
    Purdue University\\

    West Lafayette Indiana 47906 USA\\
    skose@purdue.edu, kocsis0@purdue.edu
}

\usepackage{bibentry}

\begin{document}

\maketitle

\begin{abstract}
Emerging Knowledge Tracing (KT) models, particularly deep learning and attention-based Knowledge Tracing, have shown great potential in realizing personalized learning analysis via prediction of students' future performance based on their past interactions. 
The existing methods mainly focus on immediate past interactions or individual concepts without accounting for dependencies between knowledge concept, referred as knowledge concept routes, that can be critical to advance the understanding the students' learning outcomes. To address this, in this paper, we propose an innovative attention-based  method by effectively incorporating the domain knowledge of knowledge concept routes in the given curriculum. Additionally, we leverage XES3G5M dataset, a benchmark dataset with rich auxiliary information for knowledge concept routes, to evaluate and compare the performance of our proposed method to the seven State-of-the-art (SOTA) deep learning models.
\end{abstract}

%

\section{Introduction}


In recent years, due to the growth and popularity of online learning platforms such as Massive Open Online Courses (MOOCs) \cite{wulf2014massive}, Intelligent Tutoring Systems (ITS) \cite{anderson1985intelligent}, Khan Academy, and Coursera, online learning has becomes one popular learning method. To support online learning, intelligent tutoring systems have merged as effective techniques, which focus on analyzing students' knowledge states and predicting their performance on future related tasks.

\begin{figure}[h]
\centering
\includegraphics[width=0.9\columnwidth]{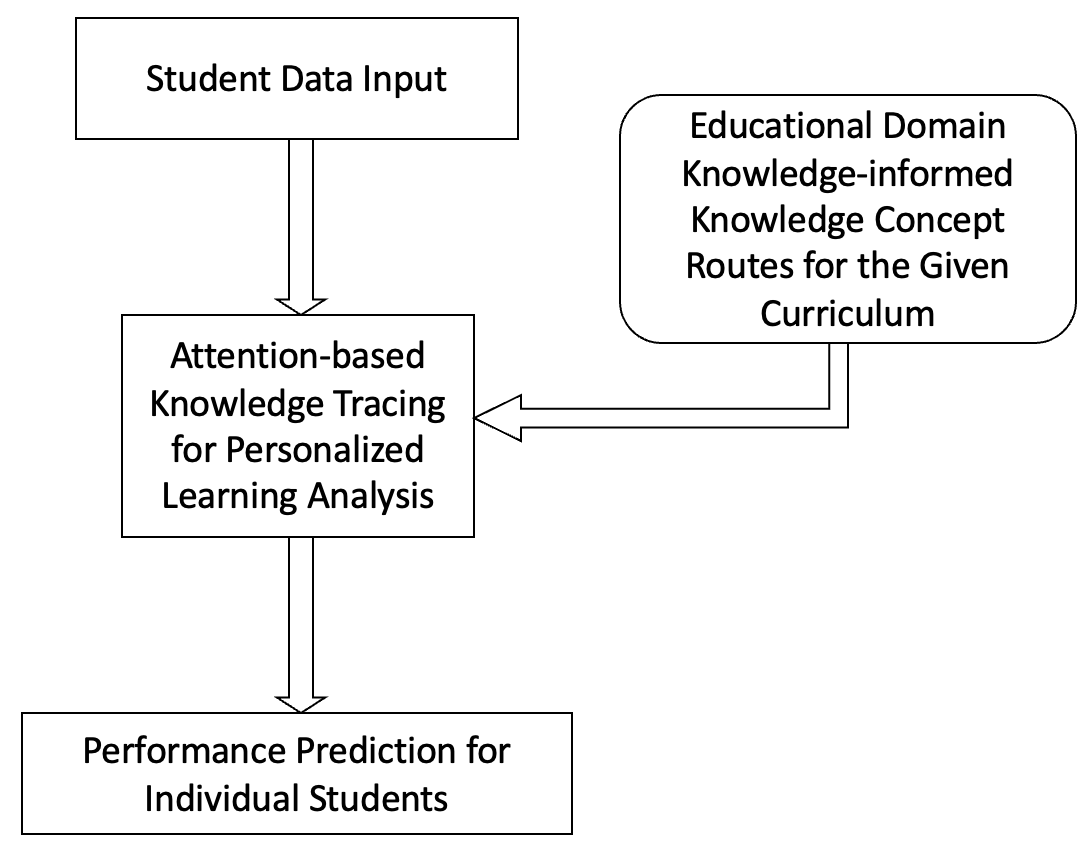} 
\caption{Overview of our Proposed Domain Knowledge-informed Attention-Based Knowledge Tracing Method}
\label{fig1}
\end{figure}

While traditional approaches, such as Bayesian Knowledge Tracing (BKT) \cite{corbett1994knowledge} and Performance Factor Analysis (PFA) \cite{pavlik2009performance} have provided foundational frameworks, these methods face limitations in capturing complex learning patterns and relationships between knowledge concepts (KCs) \cite{shen2024survey}.
Recent advances in deep learning have led to significant improvements in KT and resulted in deep learning-based KT methods like Deep Knowledge Tracing (DKT) \cite{piech2015deep} and Memory-Aware Knowledge Tracing (DKVMN) \cite{zhang2017dynamic,abdelrahman2019knowledge}. These models leverage deep learning techniques, such as Recurrent Neural Networks (RNNs), to capture temporal dependencies in student interactions. However, they have been criticized for their lack of interpretability and tendency to overfit sparse data \cite{gervet2020deep}.
In recent years, the introduction of transformers with attention mechanisms \cite{vaswani2017attention} has revolutionized deep learning across multiple domains, including natural language processing and computer vision. These models excel at capturing long-range dependencies and contextual relationships within data, making them particularly well-suited for tasks involving sequential information. This innovation has led to breakthroughs in large language models (LLMs) such as GPT-x \cite{brown2020language} and Llama \cite{touvron2023llama}, which excel at capturing long-range dependencies and contextual relationships within data. Inspired by this success, various research efforts have been conducted to develop attention-based KT methods that aim to incorporate attention mechanisms into KT models to enhance their ability to handle sparse datasets and long-term dependencies between KCs. The established methods include Self-Attentive Knowledge Tracing (SAKT) \cite{pandey2019self}, SAINT (Separated Self-AttentIve Neural Knowledge Tracing) \cite{choi2020towards} and Attentive Knowledge Tracing (AKT) \cite{ghosh2020context}. These methods enhanced the capability of traditional KT methods by focusing on ability to focus on relevant past interactions, while downweighing irrelevant ones.
However, existing studies show that these attention based KT models primarily focus on immediate past interactions or treating questions independent, posing challenges in capturing hierarchical dependencies between knowledge concepts \cite{yudelson2013individualized}. This limitation can result in suboptimal predictions, as learning process of students normally involves building upon prior knowledge in a sequential and hierarchical manner \cite{woolf2010building}.
Our work aims to address this gap by introducing a novel approach that incorporates domain knowledge-informed hierarchical knowledge concept routes of the given curriculum into the attention mechanism of KT models. By doing so, our approach is able to enhance the capability of capturing long-term dependencies between knowledge concepts and thus efficiently advance the performance. Our approach aims to improve the model’s ability to capture long-term dependencies between knowledge concepts, leading to improved predictive performance compared to existing
state-of-the-art models. To achieve this, we propose an innovative domain knowledge-informed Learning Relevance Matrix that tracks relationships between questions based on their shared concept paths. This matrix is designed to mask attention scores for unrelated questions, ensuring the model does not attend to questions involving concepts that will not aid the student in answering the current question. 
The main contributions of our work include:

\begin{itemize}
\item A novel domain knowledge-informed Learning Relevance Matrix mechanism for tracking relationships between questions based on shared knowledge concept routes.
\item An improved attention mechanism that masks attention scores for unrelated questions
\item Extensive experimental validation demonstrating significant improvements in AUC and accuracy over SOTA KT methods.

\end{itemize}
Our experiments on the benchmark dataset demonstrate that the proposed approach effectively captures hierarchical dependencies between knowledge concepts, offering improved interpretability of the underlying learning process and a promising direction for future research in personalized learning systems.

\section{Literature Review}

KT has evolved significantly over the years, with earliest model like Bayesian Knowledge Tracing (BKT) \cite{corbett1994knowledge}, that relied on probabilistic frameworks like Hidden Markov Models (HMMs) to update the probability of mastery for each KC for modelling student knowledge as a binary variable representing mastery or non-mastery of a concept. BKT assumes that once a student masters a concept, they do not forget it, which limits its ability to model real-world learning scenarios where forgetting is common \cite{baker2008more}. Logistic models such as Learning Factor Analysis (LFA) \cite{cen2006learning} and Performance Factor Analysis (PFA) \cite{pavlik2009performance} which uses logistic regression to evaluate the effects of instructional factors and learning opportunities on student performance. However, these models face challenges in handling complex sequential learning dependencies and require substantial historical data.

The introduction of deep learning models, such as Deep Knowledge Tracing (DKT), marked a significant shift by leveraging Recurrent Neural Networks (RNNs) specifically Long Short-Term Memory (LSTM) networks \cite{hochreiter1997long} to capture temporal dependencies in student interactions. Despite its huge success, DKT's lack of interpretability and inability to capture concept-specific mastery \cite{yeung2018addressing}, led to various extensions. For example, Memory-Aware Knowledge Tracing (DKVMN) introduced external memory structures to track mastery levels for individual KCs, improving both interpretability and performance \cite{zhang2017dynamic}. Recent advancements have also explored hierarchical structures within KT models. Graph-based Knowledge Tracing (GKT) \cite{nakagawa2019graph} uses Graph Neural Networks (GNNs) to model relational dependencies between KCs, improving both prediction accuracy and interpretability. However, the complexity and computational demands of these models pose a challenge for practical deployment.

Attention mechanisms have proven highly effective in addressing some of these issues. Self-Attentive Knowledge Tracing (SAKT) introduced a purely attention-based approach that focuses on relevant past interactions without relying on temporal order, improving performance on sparse datasets
\cite{pandey2019self}. However, SAKT struggles with capturing long-term dependencies between distant interactions. To address this, models such as SAINT \cite{choi2020towards} and SAINT+ \cite{shin2021saint+} incorporated deeper self-attentive layers and temporal features like elapsed time and lag time to enhance prediction accuracy by modeling the time dynamics of student interactions. To reduce the impact of noise and overfitting, ATKT \cite{guo2021enhancing} introduces an adversarial loss that helps model learn to capture more stable and reliable patterns in student responses and generate predictions that are less sensitive to small perturbations. 
Additionally, context-aware Attentive Knowledge Tracing (AKT) \cite{ghosh2020context} integrates psychometric principles with attention mechanisms to prioritize recent activities while diminishing the influence of earlier events, and at the same time giving enough weightage to the earlier events that are relevant. AKT's use of Rasch model-based embeddings accounts for question difficulty variations, further enhancing interpretability \cite{rasch1993probabilistic}. Furthermore, extraKT introduced by \cite{li2024extending} is effective where students' interaction sequences vary in length. It effectively models short-term forgetting behaviors and maintains robust predictive performance even with longer context windows.  However, it requires careful tuning, and its scalability to longer sequences remains a challenge.

Despite these advancements, existing KT models have limitation in fully capturing hierarchical dependencies between knowledge concepts. Most models focus on immediate past interactions or treat questions independently without considering how knowledge builds sequentially over time, which is highly required to provide actionable feedback and personalized learning experience to the student. This gap highlights the need for more sophisticated attention mechanisms that can track relationships between KCs across longer sequences.
Incorporating auxiliary information into KT models has also shown promise. For instance, Relation-aware Self-Attention for Knowledge Tracing (RKT) \cite{pandey2020rkt} captures relational dependencies between exercises using self-attention mechanisms, while Exercise-Aware Knowledge Tracing (EKT) \cite{liu2019ekt} embeds both student interaction data and exercise features to improve performance predictions.

In summary, while significant progress has been made in KT through deep learning and attention-based models, challenges remain in effectively capturing long-term dependencies and hierarchical relationships between KCs. This paper builds upon these advancements and address the remaining challenges of the SOTA models by innovatively incorporating domain knowledge-informed knowledge content routes in KT systems to further improve the performance of personalized learning analysis.


\section{Methodology}


\subsection{Problem Modeling}

The modeling of the KT problem is stated as: \\ For a particular learner at time step $t$, we denote their interaction with a question as a tuple $(q_t, c_t, r_t)$, where $q_t \in \mathbb{N}^+$ is the question index, $c_t \in \mathbb{N}^+$ is the concept index, and $r_t \in \{0, 1\}$ is the binary-valued response (1 for correct, 0 for incorrect). Given a learner's past history up to time $t-1$ as $\{(q_1, c_1, r_1), ..., (q_{t-1}, c_{t-1}, r_{t-1})\}$, our goal is to predict their response $r_t$ to question $q_t$ on concept $c_t$ at the current time step $t$.

\subsection{Attention Mechanism}

The attention-based models operate by computing attention scores between the current exercise \(e_{t+1}\) and all previous exercises \(\{e_1, e_2, \dots, e_t\}\). The attention mechanism assigns weights based on the relevance of each past exercise to the current one. Mathematically, the self-attention score for each pair of interactions (i,j)in the sequence is computed as \cite{vaswani2017attention}:

\[
\text{Attention}(Q, K, V) = \text{softmax}\left(\frac{QK^T}{\sqrt{d}}\right)V
\]
where \(Q = W_Q e_{t+1}\) is the query vector for the current exercise, \(K = W_K e_i\) are the key vectors for past exercises, \(V = W_V e_i\) are the value vectors for past exercises, \(W_Q\), \(W_K\), and \(W_V\) are the learned projection matrices, \(d\) is the dimensionality of the latent space.

\subsection{Attention-based KT Method}

The attention-KT method \cite{ghosh2020context} consists of four main components described below. The real-valued embedding vectors $x_t \in \mathbb{R}^D$ and $y_t \in \mathbb{R}^D$ represent each question and each question-response pair $(q_t, r_t)$, respectively.

\begin{enumerate}
    \item The \textbf{Question Encoder} outputs context-aware question embeddings:
\[\hat{x}_t = f_{enc1}(x_1, ..., x_t)\]

\item  The \textbf{Knowledge Encoder} outputs context-aware knowledge embeddings:
\[\hat{y}_{t-1} = f_{enc2}(y_1, ..., y_{t-1})\]

    \item The \textbf{Knowledge Retriever} computes the current knowledge state:

\[h_t = f_{kr}(\hat{x}_1, ..., \hat{x}_t, \hat{y}_1, ..., \hat{y}_{t-1})\]

    \item The \textbf{Response Prediction network} outputs the probability of a correct response:

\[\hat{r}_t = \sigma(f_{pred}([h_t; x_t]))\] where $\sigma$ is the sigmoid function and $f_{pred}$ is a fully-connected network.

\end{enumerate}

\subsubsection{Monotonic Attention Mechanism}

To account for temporal dynamics, the attention-based KT method uses a modified monotonic attention mechanism \cite{ghosh2020context} that introduces an exponential decay term to downweight distant past interactions. The the scaled dot-product attention values $\alpha_{t,\tau}$ are then calculated as:

\[\alpha_{t,\tau} = \frac{\exp(s_{t,\tau})}{\sum_{\tau'} \exp(s_{t,\tau'})}\] where: \[
s_{t,\tau} = \frac{\exp(-\theta \cdot d(t,\tau)) \cdot Q_t^T K_\tau}{\sqrt{d_k}}
\] \(\theta > 0\) is a learnable decay rate parameter and \(d(t,\tau)\) is a temporal distance measure between time steps \(t\) and \(\tau\).

This ensures that more recent interactions have higher influence on predicting future performance while still considering concept similarity.

\subsubsection{Context-Aware Representations}

The context-aware distance measure between the different question pair is defined as:

\[d(t,\tau) = |t - \tau| \cdot \prod_{t'=\tau+1}^t \gamma_{t,t'}\]

\[\gamma_{t,t'} = \frac{\exp(\frac{q_t^T k_{t'}}{\sqrt{D_k}})}{\sum_{1 \leq \tau' \leq t} \exp(\frac{q_t^T k_{\tau'}}{\sqrt{D_k}})}\]\\

\subsubsection{Rasch Model-Based Embeddings}

The attention-based KT method uses Rasch model-based embeddings \cite{ghosh2020context} as the raw embeddings for questions and responses:

Question embedding:
\[x_t = c_{c_t} + \mu_{q_t} \cdot d_{c_t}\]

Question-response pair embedding:
\[y_t = e_{(c_t, r_t)} + \mu_{q_t} \cdot f_{(c_t, r_t)}\] where $c_{c_t} \in \mathbb{R}^D$ is the concept embedding, $d_{c_t} \in \mathbb{R}^D$ summarizes variation in questions for the concept, $\mu_{q_t} \in \mathbb{R}$ is a scalar difficulty parameter, and $e_{(c_t, r_t)} = c_{c_t} + g_{r_t}$, where $g_1$ and $g_0$ are embeddings for correct and incorrect responses.

\subsection{Domain Knowledge-informed Knowledge Concept Route-based Self-attention Mechanism}
The essential idea of our proposed domain-knowledge-informed knowledge concept route-based self-attention mechanism is to track relationships between KCs across sequences of interactions and use this information to modify attention scores within the model. Specifically, we introduce a domain knowledge-informed Learning Relevance Matrix, (\(F_{i,j}\)), that identifies whether two questions in a sequence share the same concept route. In other words, the proposed Learning Relevance Matrix aims to characterize, whether they are part of the same hierarchical knowledge structure. To achieve this, this matrix is designed to capture the relationship between questions in an interaction sequence based on their full knowledge concept routes, ensuring that the attention mechanism focuses only on relevant related questions. The detailed workflow of our proposed attention-based KT method is presented in Algorithm 1. The main procedure of formulating and applying the proposed Learning Relevance Matrix can be summarized in three steps, which are stated below:\\

\begin{algorithm}[h!]
\caption{Proposed Domain Knowledge-Informed Attention-based Knowledge Tracing Method}
\label{alg:makt}
\textbf{Input}: Dataset with student interaction sequences $\{(q_i, r_i)\}$ \\
\textbf{Parameter}: Concept Routes auxiliary information \\
\textbf{Output}: Predicted probability for next question response $\hat{r}_t$

\begin{algorithmic}[1] 

\STATE \textbf{Initialize} Learning Relevance Matrix $\mathbf{F} \in \mathbb{R}^{n \times n}$ with zeros.

\FOR{each pair of questions $(q_i, q_j)$}
    \STATE Extract concept routes for $q_i$ and $q_j$.
    \IF {concept route of $q_i$ and $q_j$ share at least one common element}
        \STATE Set $\mathbf{F}_{i, j} = 1$ and $\mathbf{F}_{j, i} = 1$.
    \ELSE
        \STATE Set $\mathbf{F}_{i, j} = 0$ and $\mathbf{F}_{j, i} = 0$.
    \ENDIF
\ENDFOR

\STATE \textbf{Embed Input Sequences}: 
\[
\mathbf{q}_{\text{embed}} = \mathbf{Q}(q_{\text{data}}), \quad \mathbf{qa}_{\text{embed}} = \mathbf{QA}(r_{\text{data}})
\]

\FOR {each attention block}
    \STATE Compute raw attention scores:
    \[
    \alpha_{t, \tau} = \frac{\mathbf{q}_t \cdot \mathbf{k}_\tau^\top}{\sqrt{d_k}}
    \]
    \STATE Apply mask using Learning Relevance Matrix $\mathbf{F}$:
    \[
    \alpha_{t, \tau}^{\text{masked}} = \alpha_{t, \tau} \times \mathbf{F}_{t, \tau}
    \]
\ENDFOR

\STATE Compute attention outputs:
\[
\mathbf{v}_{\text{output}} = \text{Softmax}(\alpha_{t, \tau}^{\text{masked}}) \times \mathbf{V}
\]

\STATE Knowledge State Update using Transformer layers:
\[
\mathbf{h}_t = \text{TransformerLayer}(\mathbf{q}_{\text{embed}}, \mathbf{qa}_{\text{embed}}, \mathbf{v}_{\text{output}})
\]

\STATE Concatenate knowledge state with question embedding:
\[
\mathbf{o}_t = [\mathbf{h}_t; \mathbf{q}_{\text{embed}}]
\]

\STATE Predict response using a feed-forward neural network:
\[
\hat{r}_t = \sigma(\text{Linear}(\mathbf{o}_t))
\]

\STATE Calculate binary cross-entropy loss:
\[
\mathcal{L} = -\frac{1}{N} \sum_{i=1}^{N} (r_i  log(\hat r_i) + (1 - r_i) log(1 -  r_i))
\]

\end{algorithmic}
\end{algorithm}

\subsubsection{Step 1: Formulating the Learning Relevance Matrix} Let $\{(q_1, c_1), (q_2, c_2), \ldots, (q_t, c_t)\}$ be a student interaction sequence, where $q_i$ is the question and $c_i$ is the knowledge concept at time $i$. Then the \textit{knowledge concept route} can be formulated by the hierarchical structure of progression of knowledge concepts to the current concept $c_i$ that the student encounters , achieved according to the educational domain knowledge of the given curriculum. Now for each pair of questions, $q_i$ and $q_j$, where $i < j$, we compare their concept routes. If the concept route of $q_i$ share at least one common element with that of $q_j$, we set $\text{F}_{i, j} = 1$. Otherwise, $\text{F}_{i, j} = 0$. Therefore, we can formulate $\text{F}_{i, j}$ as follows:

\[
F_{i,j} = 
  \begin{cases} 
   1 & \text{if } q_i \text{ and } q_j \text{ share at least one common element,} \\
   0 & \text{otherwise}
  \end{cases}
\]


\subsubsection{Step 2: Attention Score Masking with Learning Relevance Matrix}

To incorporate the knowledge concept routes into the attention mechanism, we will modify the attention score computation. Let $\alpha_{t,\tau}$ be the attention score between the question at time $t$ and a past question at time $\tau$, the original monotonic attention score can be computed as:
    \[
    \alpha_{t, \tau} = \text{Softmax}\left(\frac{\mathbf{q}_t^\top \mathbf{k}_\tau}{\sqrt{d_k}}\right)
    \] 
The attention score between two questions is then modified by applying this Learning Relevance Matrix as a mask to the attention scores, which can be represented as:

    \[
    \alpha_{t, \tau}^{\text{masked}} = \alpha_{t, \tau} \times F_{t,\tau}
    \] 
By doing so, we are able to ensure that only scores from related questions , which is determined by the full knowledge concept route, are considered in the final attention output. By incorporating the Learning Relevance Matrix into the self-attention mechanism of our attention-based KT method, the context-aware embedding can be represented as:

    \[
    \hat{\mathbf{x}}_t = \sum_{\tau < t} \alpha_{t, \tau}^{\text{masked}} \cdot \mathbf{v}_{\tau}
    \]
Additionally, the knowledge state update can be formulated as:
    \[
    \mathbf{h}_t = f_{\text{retriever}}\left(\hat{\mathbf{x}}_1, \ldots, \hat{\mathbf{x}}_{t-1}\right)
    \]

\subsubsection{Step 3: Response Prediction} The masked attention scores are then applied in the knowledge retriever component of our attention-based KT method. The predicted probability $\hat{r}_t$ that a student answers the current question $q_t$ correctly can be computed using a feed-forward neural network that is formulated as:
\[
\hat{r}_t = \sigma(\mathbf{W}_o \cdot [\mathbf{h}_t; \mathbf{x}_t])
\]
where $\mathbf{h}_t$ is the retrieved knowledge state, $\mathbf{x}_t$ is the question embedding, and $\mathbf{W}_o$ are the learnable parameters of the output layer.
Furthermore, the model's parameters are trained by minimizing a binary cross-entropy loss function that is formulated as:

\[
L = - \sum_{i=1}^{N} \sum_{t=1}^{T} \left( r_t^i \log(\hat{r}_t^i) + (1 - r_t^i)\log(1 - \hat{r}_t^i)\right)
\] where \(r_t^i \in \{0,1\}\) refers to the ground truth of whether the student answers the question correctly or incorrectly, \(\hat{r}_t^i \in {0,1}\) is the predicted probability of whether student \(i\) answers the question correctly at time step \(t\).



\section{Performance Evaluations}
To evaluate the performance of our proposed method, we conduct experiments by using \textit{XES3G5M} dataset \cite{liu2024xes3g5m}, a comprehensive KT benchmark dataset with auxiliary information. We will compare the proposed model against various SOTA KT models, including DKT \cite{piech2015deep}, DKVMN \cite{zhang2017dynamic}, Adversarial training based Knowledge Tracing (ATKT) \cite{guo2021enhancing}, SAKT \cite{pandey2019self}, SAINT (Self AttentIve Neural Knowledge Tracing) \cite{choi2020towards}, extraKT (Extended Knowledge Tracing) \cite{li2024extending} and standard AKT with Rasch Model Embeddings \cite{ghosh2020context}. We use 5-fold cross-validation to assess model performance and tune hyperparameters using the validation set.

All models are trained using the Adam optimizer with a learning rate of 0.0001. We use a batch size of 64 and trained each model for 200 epochs. Early stopping was applied based on validation AUC to prevent overfitting. The embedding dimension \(D\) is set to 256 for all attention-based models \cite{liu2022pykt}.


\subsection{XES3G5M Dataset}

The XES3G5M dataset is a large-scale benchmark dataset for KT, collected from a real-world K-12 online math platform in China. It includes 7,652 math questions, categorized as 6,142 fill-in-the-blank and 1,510 multiple-choice, each with standardized answers, detailed solutions, and textual content for complexity analysis. It has 865 Knowledge Concepts (KCs) organized in a hierarchical tree structure, enabling modeling of prerequisite relationships and dependencies. It has a total of 5,549,635 interaction records from 18,066 students, capturing question IDs, KC IDs, binary response correctness, and timestamps.
Each question includes its KC routes, which are the paths from root to leaf in the KC hierarchy, providing insights into conceptual dependencies and progression. Additionally, personal identifiers are encrypted to ensure anonymity, preserving the integrity of the dataset while protecting user privacy.

\subsection{Dataset Preprocessing}

Data filtering is applied to the dataset to remove interactions that had missing student ID or any information from the 4-tuple interaction representation (question, knowledge components, response, timestamp). Students with less than 3 interactions in their sequence are filtered out. Additionally, 20\% of the interaction sequences are used as test set and the remaining 80\% randomly split into 5 folds: 4 for training, 1 for validation. The original question-response sequences are expanded to knowledge component level by repeating responses for questions associated with multiple knowledge components and then truncated the expanded sequences to maximum length of 200 interactions, while padding those sequences which are shorter with -1.

\subsection{Evaluation Metrics}
To evaluate the model's performance, we use the Area Under the Receiver Operating Characteristic Curve (AUC) as the primary metric as it measures the ability of the model to distinguish between correct and incorrect responses across different thresholds.We also use accuracy for measuring the percentage of correct predictions across all interactions.

\section{Evaluation Results and Discussions}

The experimental results, as summarized in Table 1, provide a comprehensive comparison between the performance of our proposed method and those of other seven SOTA KT methods by using the XES3G5M dataset. Our method significantly outperforms all baseline models in both AUC and accuracy.

\begin{table}[h]
\centering
\caption{Performance Comparison of our method with seven other SOTA KT methods}
\label{tab:results}
\begin{tabular}{|c|c|c|}
\hline
\textbf{Model}       & \textbf{AUC} & \textbf{Accuracy} \\ \hline
DKT                  & 0.835        & 84.12\%            \\ \hline

DKVMN                 & 0.828        & 83.88\%            \\ \hline

SAKT                  & 0.810        & 83.11\%            \\ \hline
SAINT                  & 0.851        & 84.54\%            \\ \hline
ATKT                  & 0.857        & 84.58\%            \\ \hline
extraKT                  & 0.861        & 85.06\%            \\ \hline
AKT                  & 0.862        & 85.01\%            \\ \hline
\textbf{Our Method}        & \textbf{0.975  } & \textbf{92.93\%}   \\ \hline
\end{tabular}
\end{table}

As shown in Table 1, DKT achieves an AUC of 0.835 and an accuracy of 84.12\%, which is lower compared with SAINT, ATKT, extraKT, AKT, and our proposed models. This is reasonable since this method lacks of ability to account for hierarchal and contextual information. DKVMN performs slightly less effectively compared to DKT, with an AUC of 0.828 and an accuracy of 83.88\%. SAINT further enhances performance with an AUC of 0.851 and accuracy of 84.54\%. ATKT improves over SAINT slightly with an AUC of 0.857 and an accuracy of 84.58\%. Amongst these existing SOTA methods, extraKT and AKT achieve better performance. The good performance of AKT illustrates the effectiveness of using Rasch model-based embedding and monotonic attention mechanism, which are also applied in our proposed method. The evaluation result also validates that our proposed method significantly outperforms all the considered SOTA approaches, achieving an AUC of 0.975 and an accuracy of 92.93\%. These results highlight the effectiveness of the proposed domain knowledge-informed knowledge concept route-based self-attention mechanism of our attention-based KT model. The substantial improvement in AUC demonstrates enhanced predictive capabilities and the higher accuracy suggests better alignment with ground truth student performance data.




\section{Conclusions and Future Work}

In this paper, we present our initial work on developing a new domain knowledge-informed attention-based KT method to effectively improve the effectiveness of predicting the learning progress of individual students. The core component of our proposed KT method is a domain knowledge concept route-based self-attention mechanism with knowledge-informed Learning Relevance Matrix, which leverages the educational domain knowledge of the given curriculum to enhance the interpretation of the correlation between the questions  and provide better insights into the learning trajectories of the individual students. The simulation results validate the significant improvement in both AUC and accuracy on predicting students' performances compared with SOTA methods.

Future work will focus on further optimizing the model and exploring its applicability across diverse datasets and learning scenarios. Additionally, incorporating interpretability features will be essential to provide actionable insights for educators and learners.



\bibliography{aaai25}

\begin{thebibliography}{30}
\providecommand{\natexlab}[1]{#1}

\bibitem[{Abdelrahman and Wang(2019)}]{abdelrahman2019knowledge}
Abdelrahman, G.; and Wang, Q. 2019.
\newblock Knowledge tracing with sequential key-value memory networks.
\newblock In \emph{Proceedings of the 42nd international ACM SIGIR conference on research and development in information retrieval}, 175--184.

\bibitem[{Anderson, Boyle, and Reiser(1985)}]{anderson1985intelligent}
Anderson, J.~R.; Boyle, C.~F.; and Reiser, B.~J. 1985.
\newblock Intelligent tutoring systems.
\newblock \emph{Science}, 228(4698): 456--462.

\bibitem[{Baker, Corbett, and Aleven(2008)}]{baker2008more}
Baker, R. S.~d.; Corbett, A.~T.; and Aleven, V. 2008.
\newblock More accurate student modeling through contextual estimation of slip and guess probabilities in bayesian knowledge tracing.
\newblock In \emph{Intelligent Tutoring Systems: 9th International Conference, ITS 2008, Montreal, Canada, June 23-27, 2008 Proceedings 9}, 406--415. Springer.

\bibitem[{Brown(2020)}]{brown2020language}
Brown, T.~B. 2020.
\newblock Language models are few-shot learners.
\newblock \emph{arXiv preprint arXiv:2005.14165}.

\bibitem[{Cen, Koedinger, and Junker(2006)}]{cen2006learning}
Cen, H.; Koedinger, K.; and Junker, B. 2006.
\newblock Learning factors analysis--a general method for cognitive model evaluation and improvement.
\newblock In \emph{International conference on intelligent tutoring systems}, 164--175. Springer.

\bibitem[{Choi et~al.(2020)Choi, Lee, Cho, Baek, Kim, Cha, Shin, Bae, and Heo}]{choi2020towards}
Choi, Y.; Lee, Y.; Cho, J.; Baek, J.; Kim, B.; Cha, Y.; Shin, D.; Bae, C.; and Heo, J. 2020.
\newblock Towards an appropriate query, key, and value computation for knowledge tracing.
\newblock In \emph{Proceedings of the seventh ACM conference on learning@ scale}, 341--344.

\bibitem[{Corbett and Anderson(1994)}]{corbett1994knowledge}
Corbett, A.~T.; and Anderson, J.~R. 1994.
\newblock Knowledge tracing: Modeling the acquisition of procedural knowledge.
\newblock \emph{User modeling and user-adapted interaction}, 4: 253--278.

\bibitem[{Gervet et~al.(2020)Gervet, Koedinger, Schneider, Mitchell et~al.}]{gervet2020deep}
Gervet, T.; Koedinger, K.; Schneider, J.; Mitchell, T.; et~al. 2020.
\newblock When is deep learning the best approach to knowledge tracing?
\newblock \emph{Journal of Educational Data Mining}, 12(3): 31--54.

\bibitem[{Ghosh, Heffernan, and Lan(2020)}]{ghosh2020context}
Ghosh, A.; Heffernan, N.; and Lan, A.~S. 2020.
\newblock Context-aware attentive knowledge tracing.
\newblock In \emph{Proceedings of the 26th ACM SIGKDD international conference on knowledge discovery \& data mining}, 2330--2339.

\bibitem[{Guo et~al.(2021)Guo, Huang, Gao, Shang, Shu, and Sun}]{guo2021enhancing}
Guo, X.; Huang, Z.; Gao, J.; Shang, M.; Shu, M.; and Sun, J. 2021.
\newblock Enhancing knowledge tracing via adversarial training.
\newblock In \emph{Proceedings of the 29th ACM International Conference on Multimedia}, 367--375.

\bibitem[{Hochreiter(1997)}]{hochreiter1997long}
Hochreiter, S. 1997.
\newblock Long Short-term Memory.
\newblock \emph{Neural Computation MIT-Press}.

\bibitem[{Li et~al.(2024)Li, Bai, Guo, Zheng, Hou, Zhan, Huang, Liu, Gao, and Luo}]{li2024extending}
Li, X.; Bai, Y.; Guo, T.; Zheng, Y.; Hou, M.; Zhan, B.; Huang, Y.; Liu, Z.; Gao, B.; and Luo, W. 2024.
\newblock Extending Context Window of Attention Based Knowledge Tracing Models via Length Extrapolation.
\newblock In \emph{ECAI 2024}, 1479--1486. IOS Press.

\bibitem[{Liu et~al.(2019)Liu, Huang, Yin, Chen, Xiong, Su, and Hu}]{liu2019ekt}
Liu, Q.; Huang, Z.; Yin, Y.; Chen, E.; Xiong, H.; Su, Y.; and Hu, G. 2019.
\newblock Ekt: Exercise-aware knowledge tracing for student performance prediction.
\newblock \emph{IEEE Transactions on Knowledge and Data Engineering}, 33(1): 100--115.

\bibitem[{Liu et~al.(2022)Liu, Liu, Chen, Huang, Tang, and Luo}]{liu2022pykt}
Liu, Z.; Liu, Q.; Chen, J.; Huang, S.; Tang, J.; and Luo, W. 2022.
\newblock pyKT: a python library to benchmark deep learning based knowledge tracing models.
\newblock \emph{Advances in Neural Information Processing Systems}, 35: 18542--18555.

\bibitem[{Liu et~al.(2024)Liu, Liu, Guo, Chen, Huang, Zhao, Tang, Luo, and Weng}]{liu2024xes3g5m}
Liu, Z.; Liu, Q.; Guo, T.; Chen, J.; Huang, S.; Zhao, X.; Tang, J.; Luo, W.; and Weng, J. 2024.
\newblock Xes3g5m: A knowledge tracing benchmark dataset with auxiliary information.
\newblock \emph{Advances in Neural Information Processing Systems}, 36.

\bibitem[{Nakagawa, Iwasawa, and Matsuo(2019)}]{nakagawa2019graph}
Nakagawa, H.; Iwasawa, Y.; and Matsuo, Y. 2019.
\newblock Graph-based knowledge tracing: modeling student proficiency using graph neural network.
\newblock In \emph{IEEE/WIC/ACM International Conference on Web Intelligence}, 156--163.

\bibitem[{Pandey and Karypis(2019)}]{pandey2019self}
Pandey, S.; and Karypis, G. 2019.
\newblock A self-attentive model for knowledge tracing.
\newblock \emph{arXiv preprint arXiv:1907.06837}.

\bibitem[{Pandey and Srivastava(2020)}]{pandey2020rkt}
Pandey, S.; and Srivastava, J. 2020.
\newblock RKT: relation-aware self-attention for knowledge tracing.
\newblock In \emph{Proceedings of the 29th ACM international conference on information \& knowledge management}, 1205--1214.

\bibitem[{Pavlik, Cen, and Koedinger(2009)}]{pavlik2009performance}
Pavlik, P.~I.; Cen, H.; and Koedinger, K.~R. 2009.
\newblock Performance factors analysis--a new alternative to knowledge tracing.
\newblock In \emph{Artificial intelligence in education}, 531--538. Ios Press.

\bibitem[{Piech et~al.(2015)Piech, Bassen, Huang, Ganguli, Sahami, Guibas, and Sohl-Dickstein}]{piech2015deep}
Piech, C.; Bassen, J.; Huang, J.; Ganguli, S.; Sahami, M.; Guibas, L.~J.; and Sohl-Dickstein, J. 2015.
\newblock Deep knowledge tracing.
\newblock \emph{Advances in neural information processing systems}, 28.

\bibitem[{Rasch(1993)}]{rasch1993probabilistic}
Rasch, G. 1993.
\newblock \emph{Probabilistic models for some intelligence and attainment tests.}
\newblock ERIC.

\bibitem[{Shen et~al.(2024)Shen, Liu, Huang, Zheng, Yin, Wang, and Chen}]{shen2024survey}
Shen, S.; Liu, Q.; Huang, Z.; Zheng, Y.; Yin, M.; Wang, M.; and Chen, E. 2024.
\newblock A survey of knowledge tracing: Models, variants, and applications.
\newblock \emph{IEEE Transactions on Learning Technologies}.

\bibitem[{Shin et~al.(2021)Shin, Shim, Yu, Lee, Kim, and Choi}]{shin2021saint+}
Shin, D.; Shim, Y.; Yu, H.; Lee, S.; Kim, B.; and Choi, Y. 2021.
\newblock Saint+: Integrating temporal features for ednet correctness prediction.
\newblock In \emph{LAK21: 11th International Learning Analytics and Knowledge Conference}, 490--496.

\bibitem[{Touvron et~al.(2023)Touvron, Lavril, Izacard, Martinet, Lachaux, Lacroix, Rozi{\`e}re, Goyal, Hambro, Azhar et~al.}]{touvron2023llama}
Touvron, H.; Lavril, T.; Izacard, G.; Martinet, X.; Lachaux, M.-A.; Lacroix, T.; Rozi{\`e}re, B.; Goyal, N.; Hambro, E.; Azhar, F.; et~al. 2023.
\newblock Llama: Open and efficient foundation language models.
\newblock \emph{arXiv preprint arXiv:2302.13971}.

\bibitem[{Vaswani(2017)}]{vaswani2017attention}
Vaswani, A. 2017.
\newblock Attention is all you need.
\newblock \emph{Advances in Neural Information Processing Systems}.

\bibitem[{Woolf(2010)}]{woolf2010building}
Woolf, B.~P. 2010.
\newblock \emph{Building intelligent interactive tutors: Student-centered strategies for revolutionizing e-learning}.
\newblock Morgan Kaufmann.

\bibitem[{Wulf et~al.(2014)Wulf, Blohm, Leimeister, and Brenner}]{wulf2014massive}
Wulf, J.; Blohm, I.; Leimeister, J.~M.; and Brenner, W. 2014.
\newblock Massive open online courses.
\newblock \emph{Business \& Information Systems Engineering}, 6: 111--114.

\bibitem[{Yeung and Yeung(2018)}]{yeung2018addressing}
Yeung, C.-K.; and Yeung, D.-Y. 2018.
\newblock Addressing two problems in deep knowledge tracing via prediction-consistent regularization.
\newblock In \emph{Proceedings of the fifth annual ACM conference on learning at scale}, 1--10.

\bibitem[{Yudelson, Koedinger, and Gordon(2013)}]{yudelson2013individualized}
Yudelson, M.~V.; Koedinger, K.~R.; and Gordon, G.~J. 2013.
\newblock Individualized bayesian knowledge tracing models.
\newblock In \emph{Artificial Intelligence in Education: 16th International Conference, AIED 2013, Memphis, TN, USA, July 9-13, 2013. Proceedings 16}, 171--180. Springer.

\bibitem[{Zhang et~al.(2017)Zhang, Shi, King, and Yeung}]{zhang2017dynamic}
Zhang, J.; Shi, X.; King, I.; and Yeung, D.-Y. 2017.
\newblock Dynamic key-value memory networks for knowledge tracing.
\newblock In \emph{Proceedings of the 26th international conference on World Wide Web}, 765--774.

\end{thebibliography}

\end{document}